\pdfoutput=1

\documentclass[11pt]{article}

\usepackage[final]{coling}

\usepackage{times}
\usepackage{latexsym}

\usepackage[T1]{fontenc}

\usepackage[utf8]{inputenc}

\usepackage{microtype}

\usepackage{inconsolata}

\usepackage{graphicx}
\usepackage{booktabs}

\usepackage{wrapfig}
\usepackage{hyperref}
\usepackage{graphicx}
\usepackage{multirow}
\usepackage{hyperref}
\usepackage{bm}
\usepackage{xcolor}

\usepackage{color}

\usepackage{stmaryrd}
\usepackage{fixltx2e}
\usepackage{xparse}
\usepackage{amssymb}
\usepackage{amsfonts}
\usepackage{soul}

\usepackage{mathtools}
\usepackage{comment}

\DeclareGraphicsExtensions{.png,.pdf}
\label{packs}

\NewDocumentCommand{\teal}{ }{\textcolor{teal}}

\NewDocumentCommand{\heng}{ mO{} }{\textcolor{red}{\textsuperscript{\textit{Heng}}\textsf{\textbf{\small[#1]}}}}

\NewDocumentCommand{\chenkai}{ mO{} }{\textcolor{blue}{\textsuperscript{\textit{Chenkai}}\textsf{\textbf{\small[#1]}}}}

\NewDocumentCommand{\revanth}{ mO{} }{\textcolor{violet}{\textsuperscript{\textit{Revanth}}\textsf{\textbf{\small[#1]}}}}

\NewDocumentCommand{\cheng}{ mO{} }{\textcolor{purple}{\textsuperscript{\textit{Cheng}}\textsf{\textbf{\small[#1]}}}}

\NewDocumentCommand{\jinning}{ mO{} }{\textcolor{teal}{\textsuperscript{\textit{Tie}}\textsf{\textbf{\small[#1]}}}}

\NewDocumentCommand{\ken}{ mO{} }{\textcolor{orange}{\textsuperscript{\textit{Ken}}\textsf{\textbf{\small[#1]}}}}

\NewDocumentCommand{\ke}{ mO{} }{\textcolor{cyan}{\textsuperscript{\textit{Ke}}\textsf{\textbf{\small[#1]}}}}

\newcommand\ct[1]{~\cite{#1}}
\newcommand\fu[1]{\footnote{\url{#1}}}
\newcommand\ctt[1]{~\citet{#1}}
\newcommand\rf[1]{~\ref{#1}}

\newcommand\nit[1]{\noindent\textbf{#1}~~}

\newcommand\blfootnote[1]{%
  \begingroup
  \renewcommand\thefootnote{}\footnote{#1}%
  \addtocounter{footnote}{-1}%
  \endgroup
}
\usepackage{amsmath,scalerel}

\DeclareMathOperator*{\concat}{\scalerel*{\Vert}{\sum}}

\title{Persona-DB: Efficient Large Language Model Personalization for Response Prediction with Collaborative Data Refinement}

\author{
Chenkai Sun\textsuperscript{\normalfont 1}, Ke Yang\textsuperscript{\normalfont 1}, Revanth Gangi Reddy\textsuperscript{\normalfont 1}, Yi R. Fung\textsuperscript{\normalfont 1} \\ \textbf{Hou Pong Chan\textsuperscript{\normalfont 1},} \textbf{Kevin Small\textsuperscript{\normalfont 2},} \textbf{ChengXiang Zhai\textsuperscript{\normalfont 1},} \textbf{Heng Ji\textsuperscript{\normalfont 1}}\\
\textsuperscript{\normalfont 1}University of Illinois Urbana-Champaign, \textsuperscript{2}Amazon \\
\texttt{\{chenkai5, czhai, hengji\}@illinois.edu}
  \\}

\begin{document}
\maketitle

\begin{abstract}

The increasing demand for personalized interactions with large language models (LLMs) calls for methodologies capable of accurately and efficiently identifying user opinions and preferences. Retrieval augmentation emerges as an effective strategy, as it can accommodate a vast number of users without the costs from fine-tuning. Existing research, however, has largely focused on enhancing the retrieval stage and devoted limited exploration toward optimizing the representation of the database, a crucial aspect for tasks such as personalization. In this work, we examine the problem from a novel angle, focusing on how data can be better represented for more data-efficient retrieval in the context of LLM customization. To tackle this challenge, we introduce Persona-DB, a simple yet effective framework consisting of a hierarchical construction process to improve generalization across task contexts and collaborative refinement to effectively bridge knowledge gaps among users. In the evaluation of response prediction, Persona-DB demonstrates superior context efficiency in maintaining accuracy with a significantly reduced retrieval size, a critical advantage in scenarios with extensive histories or limited context windows. Our experiments also indicate a marked improvement of over 10\% under cold-start scenarios, when users have extremely sparse data. Furthermore, our analysis reveals the increasing importance of collaborative knowledge as the retrieval capacity expands.\blfootnote{The code is available at \url{https://github.com/chenkaisun/Persona-DB}}

\end{abstract}

\section{Introduction}
\label{sec:intro}

The increasing demand for artificial intelligence-based services that can accurately predict and adapt to individual user preferences underscores the importance of personalization in today's digital landscape. Personalization enhances user experience, fostering trust and engagement beyond mere convenience. This need is particularly pronounced in applications involving large language models (LLMs), which serve as the foundation for a wide range of services, from personalized content moderation and medical assistance to educational platforms\ct{llm_app1,llm_app2}.

In this work, we aim to improve LLM personalization, the model's capability of customizing responses based on user background. For instance, understanding a user's penchant for concise versus detailed explanations can tailor the verbosity of responses, while knowledge of their professional domain can refine the relevance of examples provided. Specifically, we explore from the perspective of retrieval-augmentation\ct{ra_2,ra3}, which has the distinct advantage of mitigating the need for the computational burdens associated with training personalized models for a multitude of users and the magnitude of LLMs. In the retrieval-augmentation approach, relevant user-related data logs, such as historical social media posts and user descriptions, are fetched by a retriever. This information is then integrated into the instruction prompts for downstream LLMs to personalize responses. 

While resource-efficient, the efficacy of the method hinges on the quality of the underlying database. Current solutions for retrieval augmentation, however, have limited exploration in the facet and often depend largely on expansive yet shallow logs of user interactions and summaries, which can prevent the model from achieving optimal performance. Firstly, such an approach would make it difficult to capture the depth of user information, a vital component in the model performance in answering queries. Secondly, the approach discourages retrieval efficiency, since it needs to gather scattered information pieces to provide sufficient pertinent data for retrieval augmentation. For instance, a higher-level opinion, such as an attitude toward a political party, can often be a more broadly applicable and retrieval-efficient insight compared to an opinion about a specific entity within the party, where the former requires analyzing historical behaviors. Moreover, existing methods have yet to explore the potential synergies that could arise from intelligently leveraging the interconnectedness of different users' data to fill knowledge gaps.

Our paper delves into the critical research question: \textit{How do we address context efficiency in retrieval-augmented personalization}? In other words, considering the cost of model inference, how do we maintain the accuracy of the generated response by retrieving less? To address previous gaps, we propose a solution based on the premise that the construction of more structured, self-improved, and interrelated databases can enhance the LLM's ability to retrieve relevant information and to generalize. This leads us to the introduction of our framework Persona-DB, designed to encapsulate extensive user contexts and histories, fostering more accurate and individualized interactions.

The framework is underpinned by two integral components designed to enhance user databases for more generalizable inference. %
The first component involves a simple yet effective stage that involves distillation and induction on the user database to encapsulate and extend user personas such as opinions and preferences. The hierarchical construction process enables the application of learned insights across diverse task contexts.
The second component involves interconnecting relevant databases among users, drawing inspiration from the principles of collaborative filtering in recommendation systems. Recognizing that an individual's data may sometimes be insufficient to address the query (e.g., due to sparse data or domain-specific gaps), we facilitate the amalgamation of other users' data that may contain relevant interactions, based on the previous studies that users with analogous mindsets tend to exhibit similar behaviors and preferences\ct{reimer2022moral,sim_user}.
For instance, a user who participates in outdoor activities and values environmental sustainability, yet has only a few interactions related to renewable energy projects, can be matched with users of a similar environmental ethos who have engaged in discussions about renewable energy. This assists us in inferring the original user's viewpoints on solar energy initiatives, leveraging shared values to supplement the absence of interactions in that domain. 
In our work, we specifically achieve this by matching users based on shared persona keys identified in the first component and composing information pieces from the individual's database and the collaborative databases.
Our approach aims to make databases both more precise and context-efficient for downstream LLM usage, reducing the need to retrieve extensive data.

We evaluate the method on \textit{Response Forecasting for News Media}, where the downstream LLM is asked to predict users' responses to news messages, and on \textit{OpinionQA}, where the LLM predicts individual survey responses\ct{oursacl,opinionqa}. Our experimental evaluation reveals that Persona-DB consistently outperforms the baselines across various retrieval sizes, notably achieving better results when the maximal retrieval size is 10 times smaller than that of the baseline. The results indicate that our framework is advantageous when dealing with extensive user histories or limited context windows in LLMs. 
In scenarios where the user has starkly scarce data, our framework outperforms the baseline by a substantial margin (e.g., 11\% in the correlation metric), demonstrating the effectiveness of our approach to generalize under the case of new or infrequent users. In our analysis, we further explore the impact of knowledge contributed by collaborator databases, finding that its importance to overall performance escalates as the retrieval capacity grows.

\begin{figure*}
    \centering
    \includegraphics[width=1\linewidth]{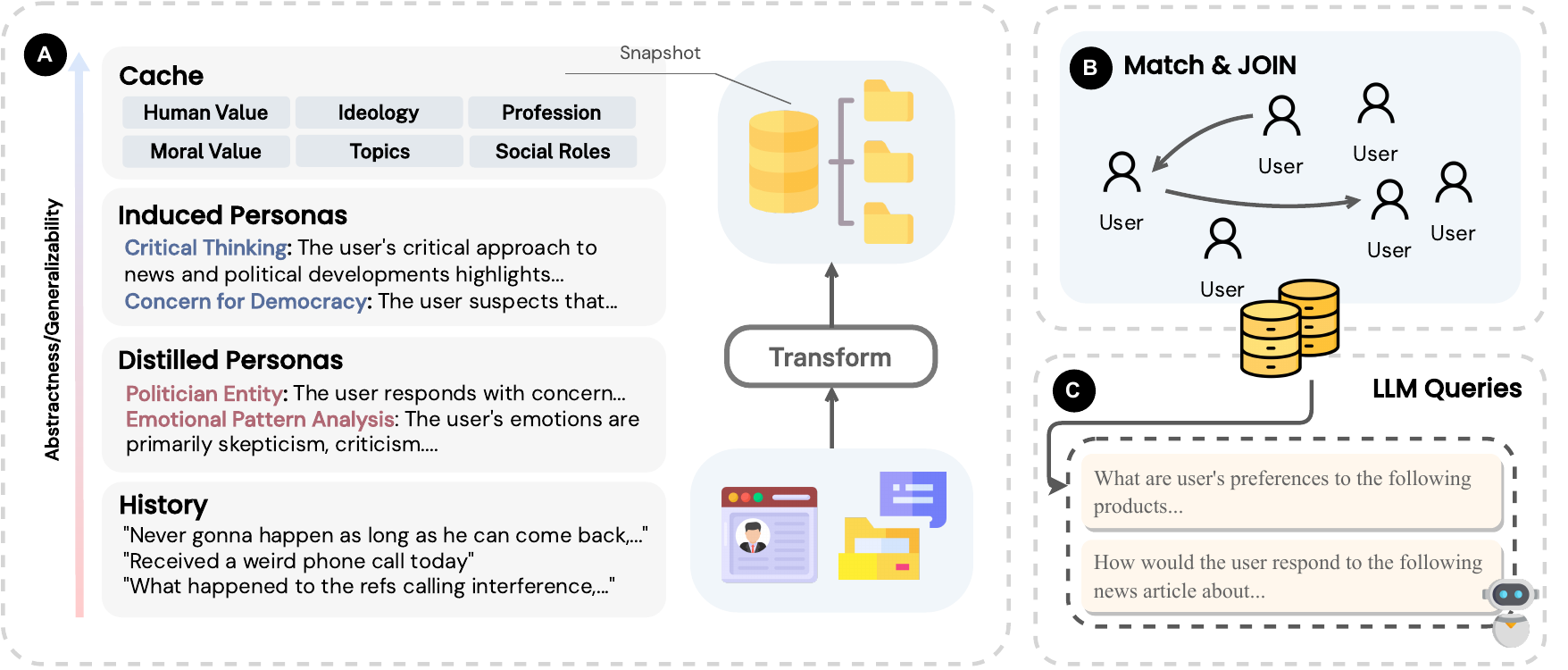}
    \caption{The image outlines the Persona-DB workflow, which starts by distilling and inducing abstract personas from users' interaction histories. It then leverages the \texttt{cache} layer to facilitate the joining of relevant user databases, effectively borrowing knowledge to fill contextual gaps in the primary user's data. This enriched data pool is subsequently used by the retrieval for personalized inference.}
    \label{fig:framework}
\end{figure*}

\section{\text{Persona-DB}}
\label{sec:method}

The primary challenge in enhancing retrieval augmentation for personalized LLM interactions is the effective and accurate representation of user context (i.e., the interaction data that inform the LLM's personalization process) within the constraints of LLM context windows and amidst the vast diversity of user behaviors. 
To effectively represent personas, it is crucial to balance comprehensiveness and data efficiency. We aim to encapsulate detailed user information, such as experiences and opinions, while ensuring that the data is efficiently retrievable. In our context, we define personalization efficiency in terms of information density and predictive power. For instance, a personalization-efficient database will be able to maintain a downstream model with fewer retrieval items.
Despite advancements in retrievers and encoders, existing methods, which predominantly rely on user logs and summaries, do not fully leverage the predictive capabilities of user personas or facilitate the exploration of relations between users.
Recognizing the limitations, we introduce Persona-DB, a novel framework designed to address key challenges in efficient user context representation. Our approach is guided by two key intuitions. Firstly, high-level (or abstract) user personas, which have been demonstrated by previous social psychology works\ct{reimer2022moral,sim_user} to effectively predict human actions across different social contexts, can provide a more generalizable basis for persona representation. Secondly, inspired by collaborative filtering prevalent in recommendation systems, we explore the potential of applying similar principles in the context of retrieval-augmented personalization. Specifically, our framework consists of two components, (1) the creation of a database by distilling and extending user data into more generalizable persona constructs and (2) a JOIN operation to bridge knowledge gaps between users by using keys identified in the previous stage. The workflow is depicted in Figure\rf{fig:framework}.

\subsection{Hierarchical Refinement}

User histories, which capture a user's past behaviors, are often noisy, with key information dispersed and sometimes only inferable through sequential analysis. Given the scenario where an LLM (or Agent) can only use a limited amount of information due to context limitations, it becomes crucial to appropriately transform the database for more efficient downstream usage.
One natural method for such transformation is creating summaries of events, yet it discourages generalization at some level. For instance, a person’s nature and beliefs (e.g.,  being supportive of an initiative) can typically be more generalizable across contexts than a reaction under a specific scenario (e.g., positive attitude toward a particular entity within the initiative). From previous works in social psychology, it was suggested that abstract personas such as values are generalizable to predicting future user actions under different social contexts\ct{reimer2022moral,ajzen1991theory}. This indicates that extracting higher-level variables from user data may introduce additional benefits to personalization accuracy while requiring lighter retrieval. To achieve this step, we choose to use an instruction-tuned LLM\ct{instr_tune,instr_tune2} to structure and infer user facts and opinions to automatically enhance the database. In our experiment, we use the same LLM for both hierarchical refinement and downstream prediction.

Our design schema is a hierarchy consisting of \texttt{History}, \texttt{Distilled Persona}, \texttt{Induced Persona}, and \texttt{Cache}, where an example snapshot is shown in Figure\rf{fig:framework}. The \texttt{History} layer contains the original user records. On the other hand, the \texttt{Distilled Persona} (abbreviated \texttt{DP}) is an induced dictionary of persona-related facts from \texttt{History}, such as superficial opinions and pattern analysis. The \texttt{Induced Persona} (abbreviated \texttt{IP}) layer comprises inferred higher-level information from the previous two components, offering a more abstract view of the user. For instance, \texttt{DP} can include specific observations, such as the user's criticism of a governmental decision to reduce public spending on community services, while \texttt{IP} can include conjectures such as that the user has a broad concern regarding social justice, derived from observations in \texttt{DP}. Lastly, \texttt{Cache} contains consistent human-defined high-level persona categories to assist relevancy matching in the collaboration stage of the framework. We utilize an LLM as an analyzer to infer and enrich the database with both user facts and opinions. In particular, each layer is populated by feeding the information from the lower layers to the LLM for processing. The structure aims to create more packed user-level features to assist data efficiency. By being able to utilize high-level, generalizable personas, the downstream model can apply these insights across a broad range of contexts, enhancing the model's ability to personalize interactions.

\subsection{Collaborative Refinement}

While using a user's data to predict intention is straightforward, it can introduce significant challenges when the user's interactions are irrelevant to a domain presented in the query or when the user simply does not have many interactions. In fact, it was discovered previously that only 25\% of highly active users generate 97\% of the content on Twitter\ct{stats_lurker}, indicating that such a case is dominant in the social media context. To address the issues, we introduce a stage to interconnect knowledge between users, under the assumption that users with similar mindsets would make similar decisions\ct{reimer2022moral}. By enabling a user to retrieve and integrate information from relevant users, we establish a collaborative database. This enriches the user context by the inclusion of potentially more domain-matching content from collaborators, thereby enhancing the generalizability and relevance of the user context representation. As demonstrated in our experiment results, such an approach indeed brings enormous benefits in cold-start situations when the user has scarce data on the platform. 

\begin{figure}[t]
	\centering
	\includegraphics[width=1\linewidth]{"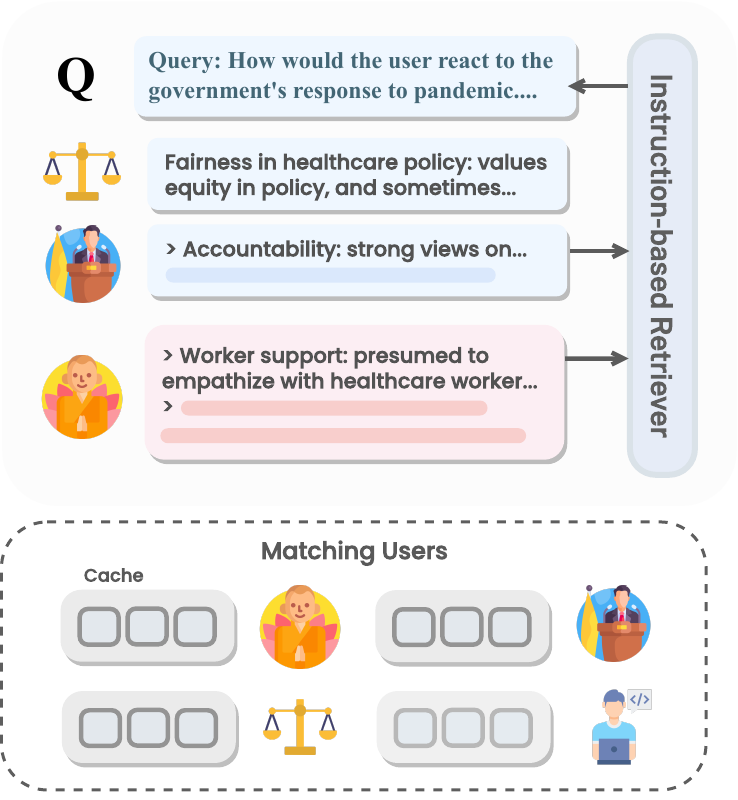"}
	\caption{During the retrieval-augmentation stage, the retriever selects data from the user's and the collaborative databases and composes them at a set ratio to inform the LLM. This strategy aims to enable the model to address challenges like sparse user interactions (e.g., cold-start) and domain irrelevance, offering an effective approach to LLM personalization in environments lacking user graphs.}
	\label{fig:stage2}
	\vskip -0.2in
\end{figure}

In our design, we name this procedure a JOIN stage as an analogy to the JOIN clause in SQL. Formally, let $\text{DB}_i$ be the persona database for user $i$, let $\text{DB}_c$ be the database for the current user, and let $\mathrm{P}_i$ indicate retrieval-based embedding prompt for user $i$. We utilize instruction-based embedding for matching in order to help the model understand the intents behind the query instead of merely comparing semantic similarity. Specifically, the process involves first encoding the \texttt{Cache} of each user's persona database using a general domain instruction-tuned $\text{LLM}_{\text{embed}}$ that has been trained on large-scale text similarity data~\cite{oneembed, ra4},
\begin{equation}
    \mathbf{X}_i = \text{LLM}_{ \text{embed}}(\mathrm{P}_i, \text{DB}_i\texttt{[Cache]}) \quad \forall i \in |U|
\end{equation}
We then use the embeddings as the key to compute similarities between the current user's database and others, followed by retrieval and joining of the most relevant databases. Formally,
\begin{equation}
    \psi(\mathbf{X}_i, \mathbf{X}_c) = \frac{\mathbf{X}_c \cdot \mathbf{X}_i}{\|\mathbf{X}_c\| \|\mathbf{X}_i\|}
\end{equation}
\begin{equation}
    \text{DB}^c_{\texttt{Join}} = \concat_{i \in \text{TopK}(\psi)}\text{DB}_i
        \label{eq:db_join}
\end{equation} 
where $\text{DB}^c_{\texttt{Join}}$ is the resulting collaborative database. During the retrieval stage, we use $\text{LLM}_{\text{embed}}$ (with a different prompt) to retrieve task-relevant information from both the database from collaborating users and the current user. We show an example of this procedure in Figure\rf{fig:stage2}. In the process of amalgamating information from the two distinct databases, we define a composition ratio denoted by $x$. This ratio indicates the allocation of the downstream retrieval capacity, $r$, such that $\lceil x\cdot r \rceil$ represents the proportion of data selected from the collaborative database, $\text{DB}^c_{\texttt{Join}}$, while the remaining capacity, $r - \lceil x\cdot r \rceil$, is sourced from the individual user database, $\text{DB}_c$. Subsequently, aggregated items are integrated into a prompt, which is then fed to a downstream LLM to produce personalized responses relevant to the user's query.

\begin{table}[]
\begin{small}
\centering
\begin{tabular}{lllll}
\toprule
 & \multicolumn{2}{c}{\textbf{$\phi_{int}$}~(\%)} & \multicolumn{2}{c}{\textbf{$\phi_{p}$}~(\%)} \\
Method & \multicolumn{1}{c}{\textcolor{purple}{$r_s$}} & \multicolumn{1}{l|}{\textcolor{purple}{$r$}} & \teal{MiF1} & \teal{MaF1} \\ \midrule
 
Majority & - & \multicolumn{1}{l|}{-} & 43.41 & 20.18 \\
Random & 0.62 & \multicolumn{1}{l|}{0.41} & 35.51 & 30.55\\

H-Retrieval & 40.96 & \multicolumn{1}{l|}{41.17} & 58.52 & 47.36 \\
H-Recency & 40.7 & \multicolumn{1}{l|}{41.49}& 58.71 & 47.49 \\
History (Full) & 42.8 & \multicolumn{1}{l|}{43.09} & 59.58 & 48.8 \\
IntSum & 44.89 & \multicolumn{1}{l|}{45.05} & 59.96 & 47.32 \\
Persona-DB & \textbf{47.67} & \multicolumn{1}{l|}{\textbf{47.88}} & \textbf{62.66} & \textbf{50.59} \\\midrule
w/o JOIN & {44.66} & \multicolumn{1}{l|}{{45.0}} & {61.79} & {50.39} \\ 
w/o IP & {43.21} & \multicolumn{1}{l|}{{43.81}} & {60.73} & {47.79} \\ 
w/o DP & {44.42} & \multicolumn{1}{l|}{{44.51}} & {60.44} & {48.72} \\

 \bottomrule
\end{tabular}
\caption{The table presents the results from the RFPN dataset, with the best overall performance highlighted in bold. Our framework (with top-40 retrieval) consistently surpasses baseline methods, notably achieving higher accuracy than the method using full user history databases. Additionally, the results include an ablation study demonstrating the standalone effectiveness of Persona-DB without the individual components.}
\label{tab:main}
\end{small}
\vskip -0.05in
\end{table}

\section{Experiment}
\label{sec:experiment}

\subsection{Dataset Overview}
\label{sec:datasets}

To achieve personalization, a model needs to be capable of predicting users' opinions and preferences regarding a wide range of potential responses or products. 
In this work, we utilize two datasets, Response Forecasting for Personas in News Media (RFPN)\ct{oursacl} and OpinionQA\ct{opinionqa}, to evaluate the model's performance in personalized response prediction. Detailed statistics are provided in the Appendix.

The RFPN dataset comprises 13.3k responses from 8.4k users to 3.8k news headlines sourced from Twitter. We utilize the dataset's test set, in which the history length can exceed 300. We evaluate the model performance on predicting the sentiment polarity $\phi_p$ (categorized as \textit{Positive}, \textit{Negative}, or \textit{Neutral}) and the ordinal intensity $\phi_{int}$ (on a scale from 0 to 3), based on a given persona and news media message. Here, a persona is constructed from user attributes, including profiles and historical posts. {}The prediction task allows for reliable evaluation, simplifying the challenge of assessing text generation in personalized conversation with lengthy user contexts, which can be challenging for both LLM and Human. The data also naturally allows the developed model to perform personalized content moderation and recommendation.
The OpinionQA dataset includes survey data from Pew Research’s American Trends Panels, and each sample consists of a single-answer multiple-choice question and the user's selection. In OpinionQA, a persona is defined by both the respondent's demographic information and historical responses. We specifically use \textit{Biomedical and Food Issues}, \textit{Global Attitudes}, and \textit{America in 2050} topics from the dataset; this allows us to assess the method's performance across diverse scenarios.

\subsection{Experimental Setup}  \label{sec:setup}

We follow the evaluation from\ctt{oursacl} for the first task. Specifically, we assess intensity predictions using Spearman and Pearson correlation coefficients, denoted as \textcolor{purple}{$r_s$} and \textcolor{purple}{$r$}, respectively. These metrics help quantify the association between the ordinal predicted and actual intensity scales, allowing for credits even when exact matches are not achieved. The sentiment polarity multi-class classification is evaluated using the traditional Micro-F1 and Macro-F1 scores, denoted \textcolor{teal}{MiF1} and \textcolor{teal}{MaF1}, respectively. For the multiple-choice-based OpinionQA task, we measure performance using accuracy. In addition to the main results, we also perform analyses of our method on the RFPN dataset. 

To test the efficacy of the Persona-DB components, we compare our framework against different variants that use the original user historical data as the database, including \textit{H-Retrieval}, where the most relevant entries are used, and \textit{H-Recency}, which uses the most recent entries. We also compare with \ctt{richardson2023integrating} (abbreviated \textit{IntSum}), which uses task-aware user summaries generated by LLMs. Additionally, we evaluate the performance of Persona-DB without collaborative refinement in a variant named Persona-DB w/o JOIN (abbreviated P-DB w/o J). For the experiments in\rf{sec:main} and\rf{sec:lurker}, we use 40 as the retrieval capacity. We also include naive strategies including \textit{Random} and \textit{Majority}, where the former makes predictions randomly, and the latter follows the majority label. A reference to using full history, \textit{History (Full)}, offers insight into the performance of complete history utilization. In the main and case studies, we set 25\% as the composition ratio.
To accommodate the context window limitation caused by an excessively large number of interactions, we limit the number of items to within the range that LLM can accept. The retrieval process utilizes an instruction-tuned LLM encoder, \texttt{text-embedding-ada-002}\fu{https://platform.openai.com/docs/models/embeddings}, to facilitate both the collaborative user retrieval and the downstream retrieval augmentation. Inference is performed using \texttt{gpt-3.5-turbo-0613}\fu{https://platform.openai.com/docs/models/gpt-3-5-turbo} via the official OpenAPI, with a fixed seed across runs. The same model is also utilized in the first stage of our framework to process user histories.

\subsection{General Case Results Discussion} %
\label{sec:main}

We evaluate the effectiveness of the Persona-DB framework and the baselines, setting the top-k as 40 for both our variants and the baselines in the retrieval-augmentation step. From the evaluation on RFPN (Table\rf{tab:main}), it is evident that H-Recency and H-Retrieval underperform compared to History (Full), as expected, due to their less comprehensive view of the user's information. On the other hand, Persona-DB consistently outperforms these retrieval-based methods across all evaluation metrics with a large margin in correlation, demonstrating that our framework can achieve superior performance under the same retrieval capacity, compared to using historical posts only. The table also indicates that while leveraging a user's complete history offers a broad view of user data during inference, our approach using only the top-k personas, in fact, exceeds the performance of History. This suggests that our strategy of self-improving and composing interrelated personas achieves better personalization efficiency as it uses much less data while achieving better results.

To underscore the significance of the collaborative stage within our framework further, we examine their absence's effect on performance. Omitting JOIN operations results in a marked decrease in correlation metrics and F1 scores, affirming its essential role in addressing knowledge gaps among users. Interestingly, even without implementing JOIN, the hierarchical database exhibits greater predictive capability than that of the baselines using a log-only database. This observation reiterates the value of high-level insights in improving generalizability in the downstream models. We also observe that removing the intermediate persona layers (i.e. IP or DP), leads to a decrease in performance, underscoring their importance in enhancing the model's capability. Additionally, we conduct a validation study to assess the quality of the LLM analyzer in stage 1 (detailed in the Appendix). Although the extraction is not perfect due to the use of LLM, it still allows both components of our framework to work. This indicates that the method is robust against error propagation at some level, and further enhancements to the LLM to generate higher-quality insights may yield improved performance for Persona-DB.

We also present results on OpinionQA in Figure\rf{fig:qa}. The plot reveals a similar pattern where Persona-DB surpasses the baseline models across topics. Notably, we see that compared with the baselines, it performs particularly well in the \textit{Global Altitudes} category, consisting of questions on international affairs. This enhanced performance can be attributed to the importance of opinions from users within similar demographic groups in predicting current users' opinions on this topic.

\begin{figure}
	\centering
	\includegraphics[width=0.9\linewidth]{"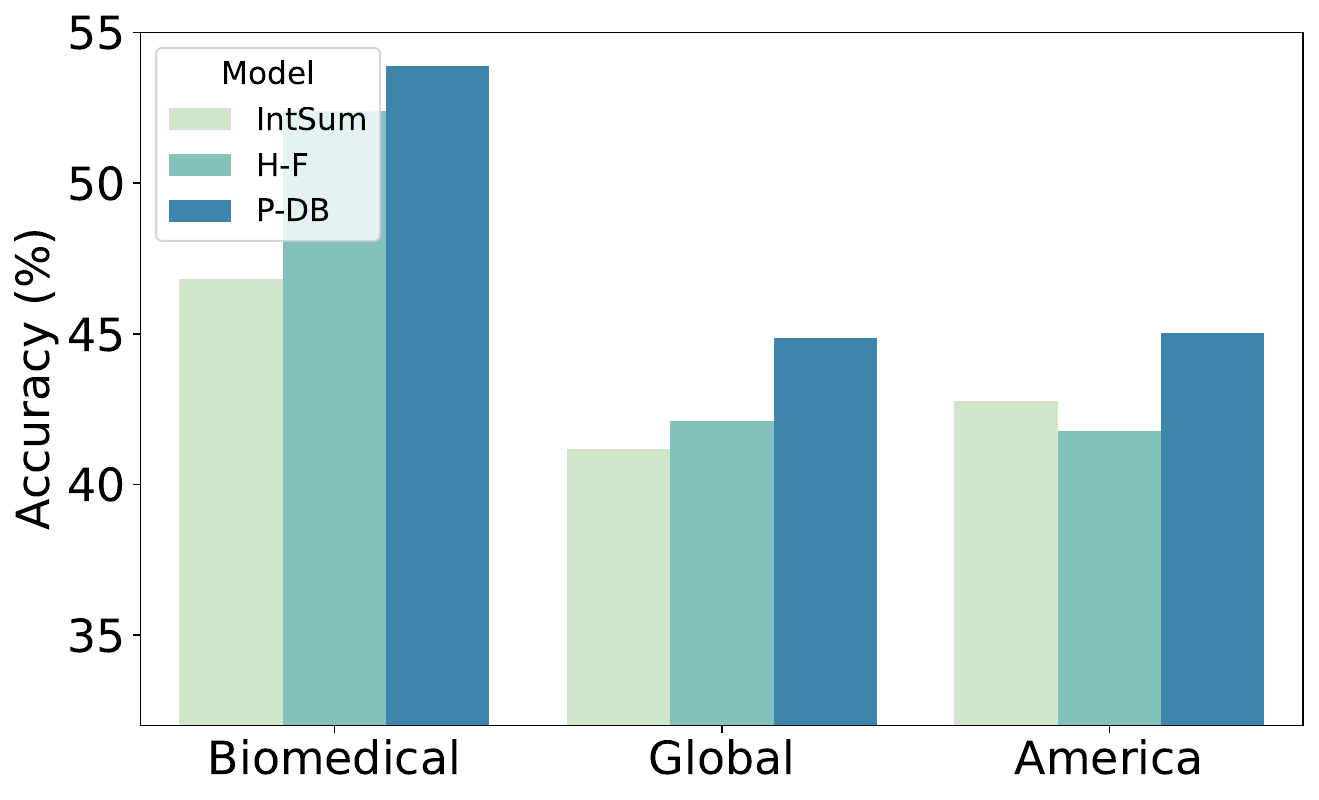"}
	\caption{Performance comparison on the OpinionQA task. The plot shows that Persona-DB outperforms the baselines consistently.}
	\label{fig:qa}
	\vskip -0.1in
\end{figure}

\begin{table}[]
\begin{small}
\centering
\begin{tabular}{lllll}
\toprule
 & \multicolumn{2}{c}{\textbf{$\phi_{int}$}~(\%)} & \multicolumn{2}{c}{\textbf{$\phi_{p}$}~(\%)} \\
Method & \multicolumn{1}{c}{\textcolor{purple}{$r_s$}} & \multicolumn{1}{c|}{\textcolor{purple}{$r$}} & \teal{MiF1} & \teal{MaF1} \\ \midrule
\multicolumn{5}{c}{Lurkers (Cold-Start)}\\
\midrule

History (Full) & 40.66 & \multicolumn{1}{l|}{41.62} & 67.96 & 46.02 \\
IntSum & 43.15 & \multicolumn{1}{l|}{45.41} & 69.9 & 46.05 \\
Persona-DB & \textbf{55.05} & \multicolumn{1}{l|}{\textbf{56.75}}& \textbf{77.67} & \textbf{57.29} \\\midrule

w/o JOIN & {47.07} & \multicolumn{1}{l|}{{53.34}} & {73.79} & {49.79} \\
\midrule
\multicolumn{5}{c}{Frequent Users}\\ 
\midrule

H-Retrieval & 30.63 & \multicolumn{1}{l|}{33.34} & 56.08 & 42.02 \\
H-Recency & 38.74 & \multicolumn{1}{l|}{40.95} & 58.75 & 46.21 \\
IntSum & 44.65& \multicolumn{1}{l|}{46.58} & 60.83 &44.67 \\
History (Full) & {43.29} & \multicolumn{1}{l|}{{45.07}} & 61.72 & 48.63 \\
Persona-DB & \textbf{49.03} & \multicolumn{1}{l|}{\textbf{49.71}} & \textbf{65.58} & \textbf{52.72} \\\midrule
w/o JOIN & 41.41 & \multicolumn{1}{l|}{44.44 }& {63.5} & {49.67} \\

 \bottomrule
\end{tabular}
\caption{The two case studies, focusing on users with extremely scarce (Lurkers) and long (Frequent Users) data histories, demonstrate that our framework markedly improves performance for both ends of the spectrum. These results underscore the effectiveness of our methods in cold-start scenarios, and the ability to process and personalize content for users with extensive histories.}
\label{tab:lurker}
\end{small}

\end{table}

\subsection{Analysis on Lurkers and Frequent Users}
\label{sec:lurker}

In our evaluation, we delve into the evaluation of Persona-DB across two types of common yet challenging cases: Lurkers and Frequent Users. This segmentation allows us to understand the framework's robustness and performance in scenarios representing the extremes of user interaction volumes. We perform the analysis on the RFPN task as it features high diversity in the history lengths.

\nit{Lurkers}
The term "Lurker" refers to users with minimal interaction history, a common scenario in recommendation systems known as the cold-start problem. Being capable of personalizing responses for lurkers indicates a framework's ability to offer tailored interactions to new users without the prerequisite of extensive historical accumulation. We select 100 users with the sparsest (nonempty) interaction records, averaging only 13.81 interactions per user. The results, as illustrated in Table\rf{tab:lurker}, highlight Persona-DB's superior performance across all metrics compared to the baselines. Specifically, our framework shows a 11\% improvement in Pearson correlation over the baseline method, showcasing its exceptional ability to maintain accuracy with minimal user data. This efficiency can be attributed to the framework's capability to discern user characteristics from limited interactions, leveraging reliable, frequent collaborators' insights to enhance personalization accuracy. Remarkably, even before applying the JOIN operation, the transformed database exhibits noticeably improved performance, underscoring the predictive power of LLMs in identifying generalizable personas.

\nit{Frequent Users}
Furthermore, we assess the framework's performance with frequent users, characterized by voluminous interaction histories, potentially introducing noise to the retrieval process. For this analysis, we focused on the top 300 users with the longest histories. The evaluation is detailed in Table\rf{tab:lurker}. Interestingly, the performance of H-Retrieval is notably lower than that of the recency-based method. One hypothesis for this observation is that when the user history becomes frequent, the retrieval of relevant information in a small capacity becomes harder as there are more semantically similar yet non-relevant items. The table also demonstrates Persona-DB's consistent outperformance over alternative methods. This finding emphasizes the framework's adeptness at tailoring content for users with extensive interaction histories. The success can be largely attributable to the hierarchical representation's ability to synthesize and distill relevant information from a broad dataset, effectively mitigating noise and enhancing personalization accuracy for both frequent users and their collaborators.

\begin{table}[]
\begin{small}
\centering
\begin{tabular}{clcc}
\toprule
Top-K & Method & Pearson & Accuracy \\ \midrule
40 & Persona-DB & \textbf{47.88} & \textbf{62.66} \\
40 & IntSum & 45.05 & 59.96 \\
40 & Hist-Recency & 41.49 & 58.71 \\
30 & Persona-DB & \textbf{47.45} & \textbf{61.89} \\
30 & IntSum & 41.93 & 59.38 \\
30 & Hist-Recency & 42.08 & 59.58 \\
10 & Persona-DB & \textbf{45.46} & \textbf{60.35} \\
10 & IntSum & 45.33 & 60.15 \\
10 & Hist-Recency & 42.21 & 59.19 \\
\bottomrule
\end{tabular}
\caption{Our framework is capable of maintaining performance above that of baselines, even when the retrieval size is significantly reduced.}
\label{tab:robust}
\end{small}
\vskip -0.05in
\end{table}

\subsection{Analysis on Varying Retrieval Sizes}

To minimize context window cost in the inference stage, it is crucial to improve data efficiency in retrieval (i.e., the retrieved set remains generalizable while being reduced in size). This section explores how varying retrieval sizes affect our framework's performance.

Table\rf{tab:robust} demonstrates the robustness of our method, illustrating that \textit{accuracy is maintained above the baselines even as the retrieval cap is significantly reduced}. This underscores the method's resilience to variations in context window size for LLMs, highlighting its capability to deliver precise predictions with minimal data retrieval. The compelling performance, even with a constrained database, supports our hypothesis that the personas derived from our hierarchical construction possess enhanced predictive power. Moreover, we can see that as the retrieval size decreases, the advantage of using collaborative entries shrinks. This trend likely stems from the increased significance of a user's own data in situations where retrieval capacity is limited.

\subsection{The Impact of Collaborative Composition}
\label{sec:impact}

We further analyze how the proportion of collaborative items influences Pearson correlation performance across different retrieval capacities. The heatmap depicted in Figure\rf{fig:comp} reveals an interesting trend: for larger retrieval sizes, incorporating a higher percentage of data from collaborative databases results in improved performance. Conversely, for smaller retrieval sizes, an increased reliance on collaborative data sometimes detrimentally affects performance. This pattern can be attributed to the reason that typically, a user's database contains only a limited number of entries directly relevant to addressing a query effectively. Once the retrieval size surpasses this small pool of pertinent data, the inclusion of novel, domain-relevant insights from collaborative databases becomes increasingly beneficial. However, in situations with extremely limited retrieval capacity, the critical, predictive nature of the user's own data takes precedence. In these instances, over-reliance on external data sources can, therefore, undermine performance. The pattern underscores the importance of maintaining a strategic balance in the composition of the retrieval set to optimize LLM personalization.

\begin{figure}
	\centering
	\includegraphics[width=0.9\linewidth]{"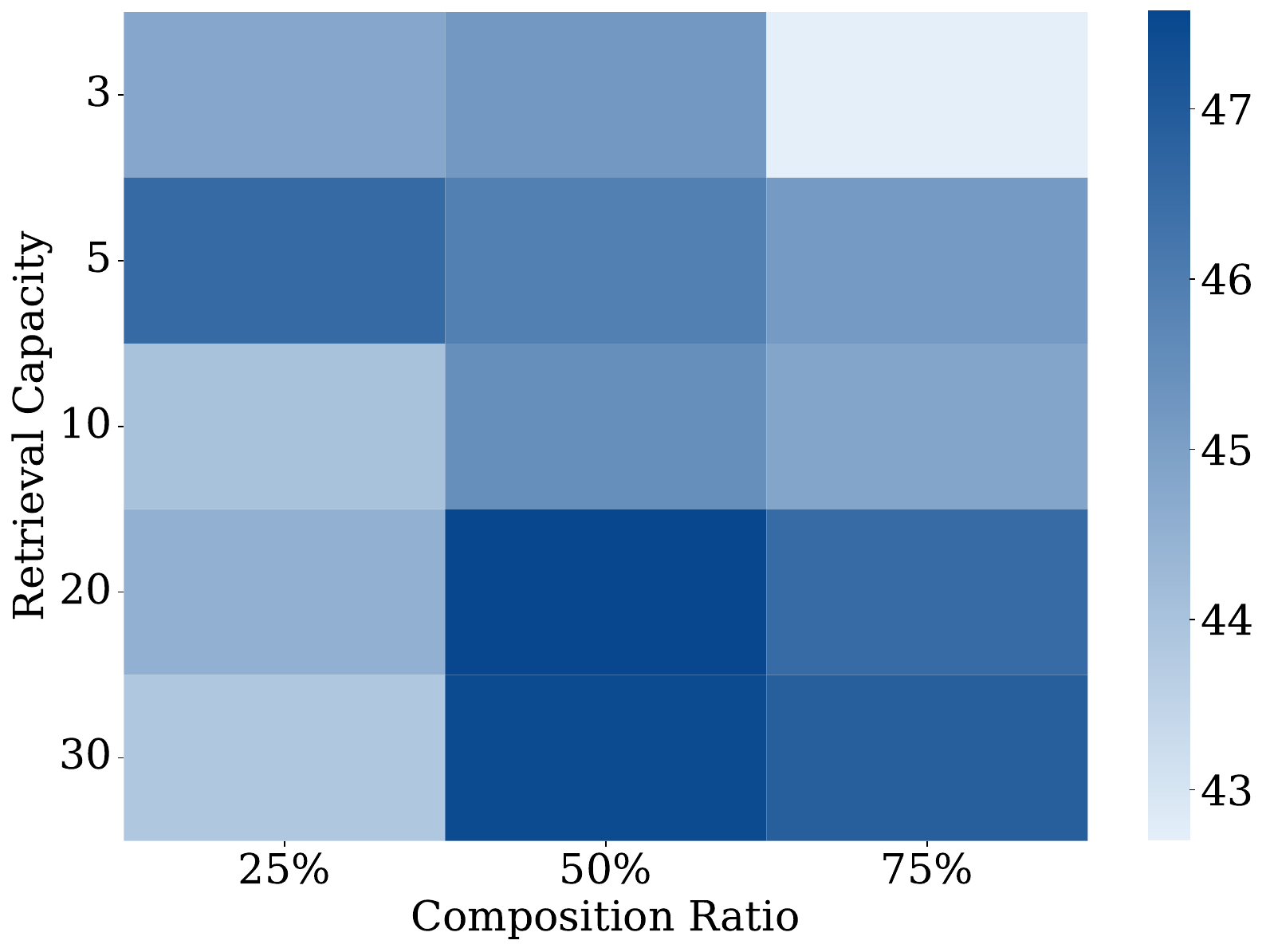"}
	\caption{The figure illustrates the shift of correlation performance metric as capacity and proportion of collaborative content changes. The trends show that collaborative retrieval becomes more important as the retrieval size grows.}
	\label{fig:comp}
	\vskip -0.2in
\end{figure}

\section{Related Work}
\label{sec:related}

Personalization plays a crucial role in the effectiveness of large language models (LLMs), particularly in ensuring their alignment with individual preferences and interests\ct{Hwang2023alignuseropinion,related_llm1,related_llm2,lmswitch,controlsum,ge2024scaling,salemi2023lamp}. With the advent of LLMs, the landscape of personalization has also been significantly transformed, enabling enhanced user engagement and the delivery of services tailored to individual needs\ct{related_llm3} and cultural preferences\ct{li-etal-2023-defining,fung-etal-2023-normsage,fung2024massively,DBLP:journals/corr/seallms3}. Traditional personalization approaches for language models often use auxiliary encoders to learn user representation from user profile data\ct{tradition1,tradition3,tradition2,steerability}. For instance, to personalize pre-trained language models, UserIdenitifer~\cite{DBLP:conf/naacl/MireshghallahSS22} prepends a user identifier prefix at the beginning of the input text during the fine-tuning process. Yet such works would require training resources and can be challenging in situations of extensive user data and large models. Recently, there have been works focusing on utilizing and improving retrieval augmentation due to its capability to efficiently borrow external knowledge lacking in the pre-trained data\ct{ra_1,ra_2} without the cost of fine-tuning downstream LLMs. In\ctt{ra3} and\ctt{ra4}, the authors introduced new embedding models to better support the diverse retrieval augmentation needs. Furthermore,\ctt{ra5} introduced a practical approach for distilling retrieval-augmented LLMs, and\ctt{ra6} demonstrated the effectiveness of retrieval-augmented language model pre-training for open-domain question answering. Yet limited exploration has been done on optimizing the data to be retrieved to improve personalization.

\section{Conclusions and Future Work}
\label{sec:conclusion}

In conclusion, we introduce Persona-DB, a framework designed to enhance the accuracy and context efficiency of retrieval-based LLM personalization through collaborative data refinement. The framework leverages a hierarchical approach to user data representation, enabling the construction of insights that are generalizable across task contexts, and introduces a collaborative mechanism to effectively bridge communal knowledge for improved query response when the user lacks sufficient relevant data. Our experimental results indicate a marked performance improvement, particularly in scenarios characterized by sparse user data, thus addressing a critical challenge in LLM personalization. Furthermore, we perform analysis to study the impact of data composition on retrieval effectiveness. Future work will focus on developing methods that refine the framework's capability to dynamically adapt the matching process based on user feedback and interaction patterns.

\section*{Limitations}
\label{sec:limitation}

We employ LLMs for abstracting personas in our work. Due to inherent bias and imperfection in the LLM, the quality of the extraction in hierarchical refinement can also affect the generalizability on diverse personas. We believe that this shortfall can be addressed through continuous improvement of the LLM's capabilities. Moreover, in this study, our focus primarily lies on improving retrieval efficiency in downstream task scenarios, where we evaluate the model's capability to preserve accuracy while reducing retrieval capacity, which in turn reduces the cost associated with online inference. In the hierarchical refinement stage, there is an existence of a one-time inference cost associated with preprocessing user data, a trade-off for enhancing data efficiency in downstream usage. We believe that using a smaller distilled model will alleviate the cost for this stage.

\section*{Ethics Statements}
\label{sec:ethics}

Our personalization approach involves storing and retrieving information about user preferences and personalities, which requires careful safety protocol and ethical consideration when deployed. While the extraction of generalized user personas and collaborative data composition offers powerful personalization capabilities, the sharing and combining of information between users requires consideration of privacy implications. The process should be implemented with appropriate privacy protections such as data anonymization, secure storage, and user consent.

Additionally, we recognize important considerations for responsible development: First, personas can risk oversimplifying and stereotyping users (as they are caricatures of people), potentially leading to biased or unrealistic representations that are unable to capture the true complexity and diversity of human behavior. Second, personalization technology enables detailed user profiling that could be misused for surveillance or targeting of individuals and groups. To mitigate these risks, the developers need to emphasize privacy-preserving techniques and aim to maintain user agency by being transparent about data usage.

Moreover, the datasets we used in this work are from publicly accessible repositories from existing publications. We only provide pointers to these existing data repositories and the data loading script that contains no user information. We will not share the datasets themselves. We will also not release any intermediate representations generated by models which may contain user information.

\section*{Acknowledgement} This research is based upon work supported in part by U.S. DARPA INCAS Program No. HR001121C0165. The views and conclusions contained herein are those of the authors and should not be interpreted as necessarily representing the official policies, either expressed or implied, of DARPA, or the U.S. Government. The U.S. Government is authorized to reproduce and distribute reprints for governmental purposes notwithstanding any copyright annotation therein. 

\bibliography{coling_latex}
\newpage
\appendix

\section{Appendix}
\label{sec:appendix}

\subsection{Additional Details}
\label{app:training}
Our experiments were conducted using the PyTorch framework~\cite{pytorch} and the Huggingface Transformers library~\cite{huggingface}. The sentiment intensity labels adhere to the definition in SemEval-2018 Task 1\ct{mohammad2018semeval}, incorporating both magnitude and direction in the evaluation of sentiment intensity. In our work, we use \texttt{concatenation} of histories to implement the JOIN in Equation\rf{eq:db_join} after collaborative databases have been retrieved. Future work will explore condensing the collaborative databases using LLMs.

The data statistics for RFPN are shown in Table\rf{tab:data_stats}. For OpinionQA, each topic we used contains a test split of 1000 samples. In particular, \textit{Biomedical and Food Issues} contains 67 questions in the survey and there are 2537 respondents in total, \textit{Global Attitudes} contains 104 questions and 2596 respondents, and \textit{America in 2050} contains 90 questions with 2524 respondents in total.

We show all prompts used in the work in the following figures. Figure\rf{fig:e1} and Figure\rf{fig:e2} represent the prompts for constructing the hierarchical database. Figure\rf{fig:p1}, Figure\rf{fig:p2}, and Figure\rf{fig:p3} represent the inference prompt for baseline, Persona-DB w/o JOIN, and Persona-DB.

\subsection{Validation Study on LLM Extraction}
\label{app:vs}

As the extracted personas in stage 1 serve as an intermediate layer to the method, and the LLM analyzer itself can introduce errors, we conducted a validation study regarding the quality of the extraction. Specifically, we sampled 50 LLM extraction results from user histories and distributed them to three human raters (who are graduate students who passed an initial quiz of 8 samples) to verify the accuracy. On average, the raters assigned a score of 3.9/5, indicating a notable level of precision despite imperfections in the LLM's output. While the extraction accuracy isn’t imperfect, our results show that the quality of the current annotation does have the merit of contributing to performance positively.

\subsection{Comparison with Prompt Compression}
\label{app:lc}

In our method, we focus on enhancing data efficiency through retrieval augmentation. Another line of approach involves compressing the prompt. Although prompt compression methods can be integrated with our approach, either before or after, it would be beneficial to conduct a comparison using the datasets in our study. Specifically, we compare our method with LLMLingua-2\ct{llmlingua2} with a 40\% compression rate on the full history (e.g., the actual number of tokens produced can vary due to probabilistic inference). LLMLingua-2 employs a data distillation procedure to extract knowledge from an LLM for prompt compression. The comparative results are presented in Table\rf{tab:lc_comp} and Figure\rf{fig:lc_comp}. Our initial analysis indicates that when comparing the two independent directions of methods separately, our method outperforms in the current setting; this may stem from our method's integration of knowledge from collaborative users.

{\renewcommand{\arraystretch}{1} 
\begin{table}[]
\begin{small}
	\centering
    \begin{tabular}{l|ccc}
   \toprule
Split & Train & Dev. & Test \\\midrule
\# Samples & 10,977  & 1,341 & 1,039 \\
\# Headlines & 3,561 & 1,065 & 843 \\
\# Users & 7,243 & 1,206  & 961 \\
Avg \# Profile Tokens &10.75  & 11.02 & 10.50 \\
Avg \# Response Tokens & 12.33 & 12.2 & 11.87 \\
Avg \# Headline Tokens &19.79  &  19.82& 19.72 \\\bottomrule 
\end{tabular}
    
    \caption{Summary statistics for the RFPN dataset.}
	\label{tab:data_stats}
\end{small}
\end{table}
}

\subsection{Additional Evaluation Metrics and Qualitative Results}
\label{app:aditional_metrics}
We show results including 1-Wasserstein distance-based alignment score and mean squared error metrics for sentiment intensity prediction in RFPN (Table\rf{tab:aditional_metrics}). We see that the results match the trend in the main table.

We additionally show qualitative case studies in Figure\rf{fig:cases} to demonstrate the benefits of collaborative strategy.

\begin{table}[]
\begin{small}
\centering
\begin{tabular}{lll}
\toprule
\textbf{method} & \textbf{wasserstein} & \textbf{mse} \\
\midrule
H-Retrieval    & 72.77 & 4.45 \\
H-Recency      & 72.71 & 4.44 \\
IntSum      & 73.15 & 4.3 \\
History (Full) & 73.46 & 4.3 \\
Persona-DB     & \textbf{74.27} & \textbf{4.13} \\\midrule
w/o JOIN       & 73.13 & 4.32 \\
w/o IP         & 72.88 & 4.49 \\
w/o DP         & 72.98 & 4.4 \\
\bottomrule
\end{tabular}
\caption{Results on additional metrics.}
\label{tab:aditional_metrics}
\end{small}
\end{table}

\begin{figure*}
	\centering
	\includegraphics[width=0.9\linewidth]{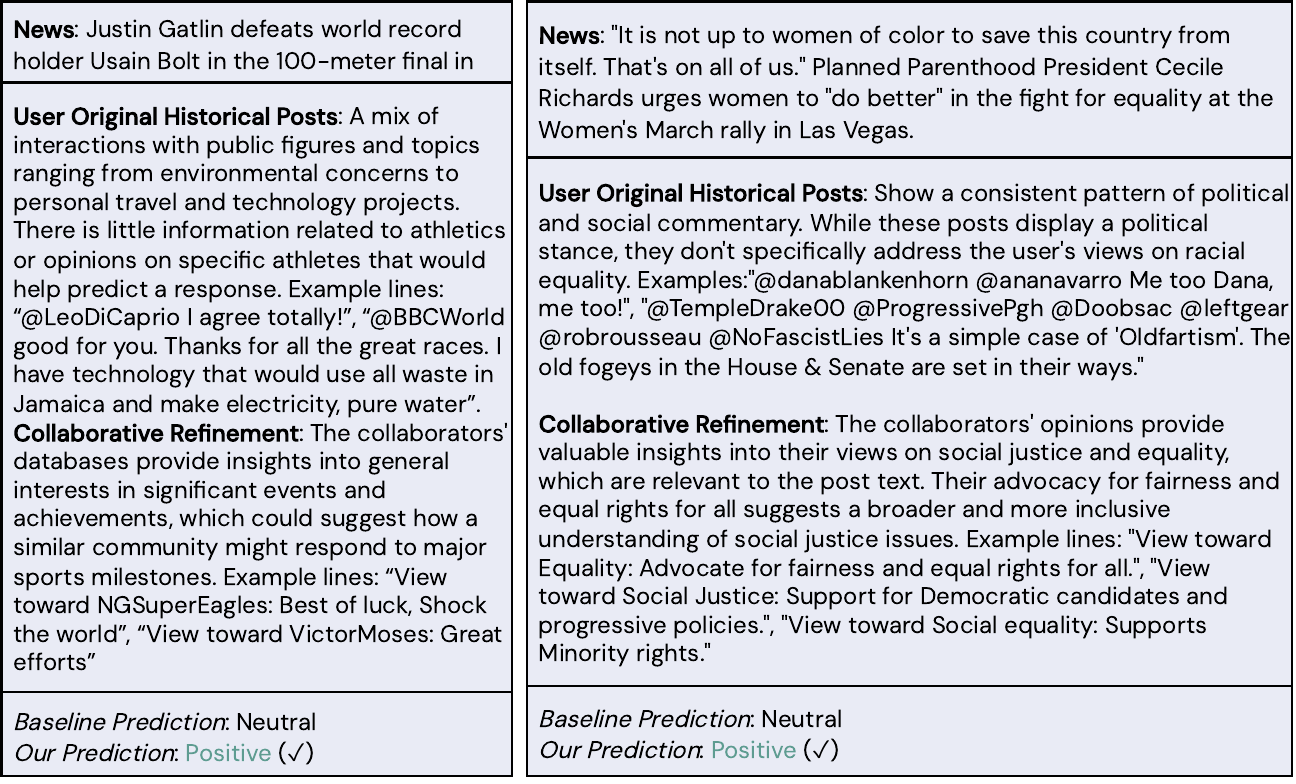}
	\caption{Case Studies.}	
    \label{fig:cases}
\end{figure*}

\begin{table}[t]
\vskip -1.75in
\begin{small}
\centering
\begin{tabular}{lllll}
\toprule
 & \multicolumn{2}{c}{\textbf{$\phi_{int}$}~(\%)} & \multicolumn{2}{c}{\textbf{$\phi_{p}$}~(\%)} \\
Method & \multicolumn{1}{c}{\textcolor{purple}{$r_s$}} & \multicolumn{1}{l|}{\textcolor{purple}{$r$}} & \teal{MiF1} & \teal{MaF1} \\ \midrule

LLMLingua-2 & 42.46 & \multicolumn{1}{l|}{42.94} & 59.96 & 48.87 \\
Persona-DB & \textbf{47.67} & \multicolumn{1}{l|}{\textbf{47.88}} & \textbf{62.66} & \textbf{50.59} \\

 \bottomrule
\end{tabular}
\caption{The table displays the RFPN task results when compared to prompt compression.}
\label{tab:lc_comp}
\end{small}
\vskip -2.05in
\end{table}

\begin{figure}[t]
\vskip -1.75in
	\centering
	\includegraphics[width=0.85\linewidth]{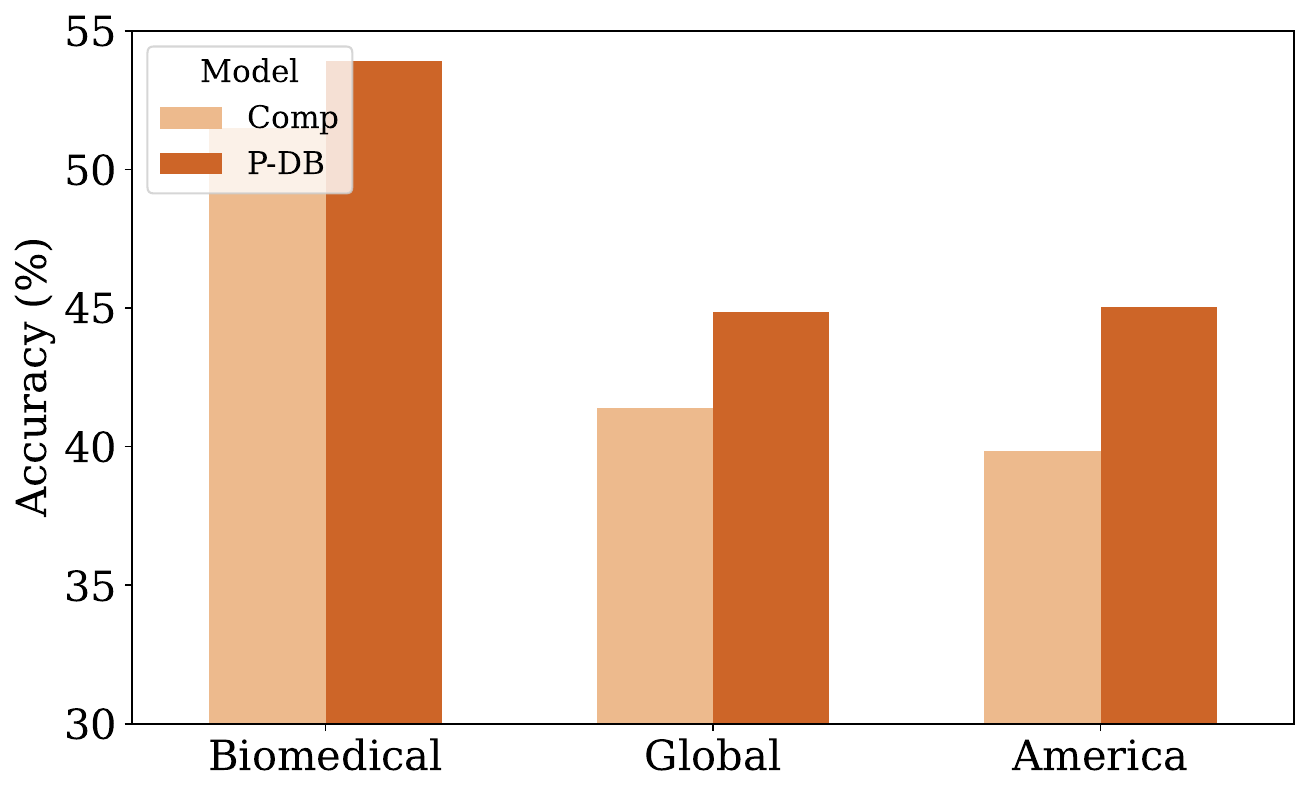}
	\caption{Comparison with prompt compression on OpinionQA task.}	
    \label{fig:lc_comp}
\end{figure}

\begin{figure*}
	\centering
	\includegraphics[width=1\linewidth]{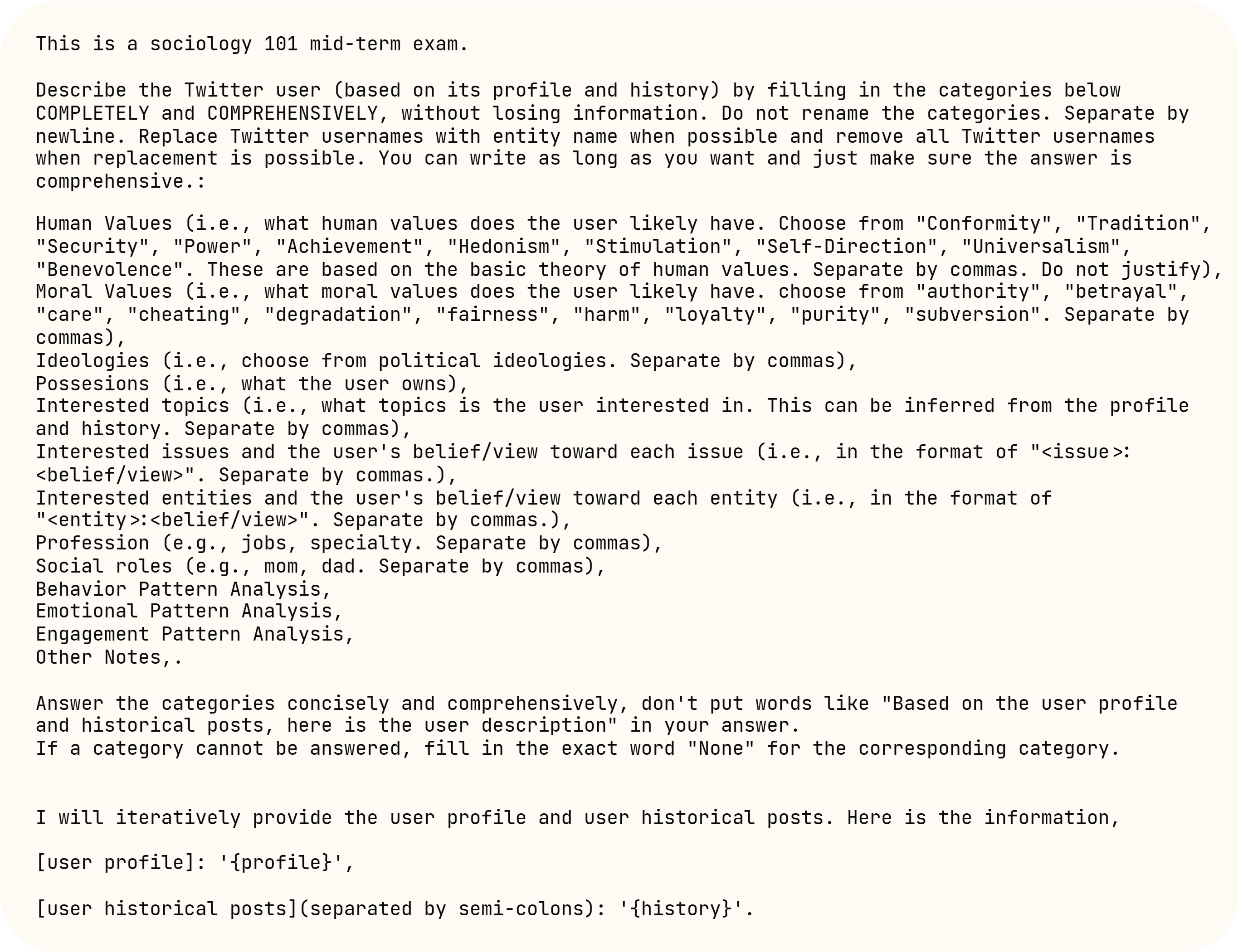}
	\caption{Prompt template used for extracting hierarchical user database.}
	\label{fig:e1}
	\vskip -0.3in
\end{figure*}

\begin{figure*}
	\centering
	\includegraphics[width=1\linewidth]{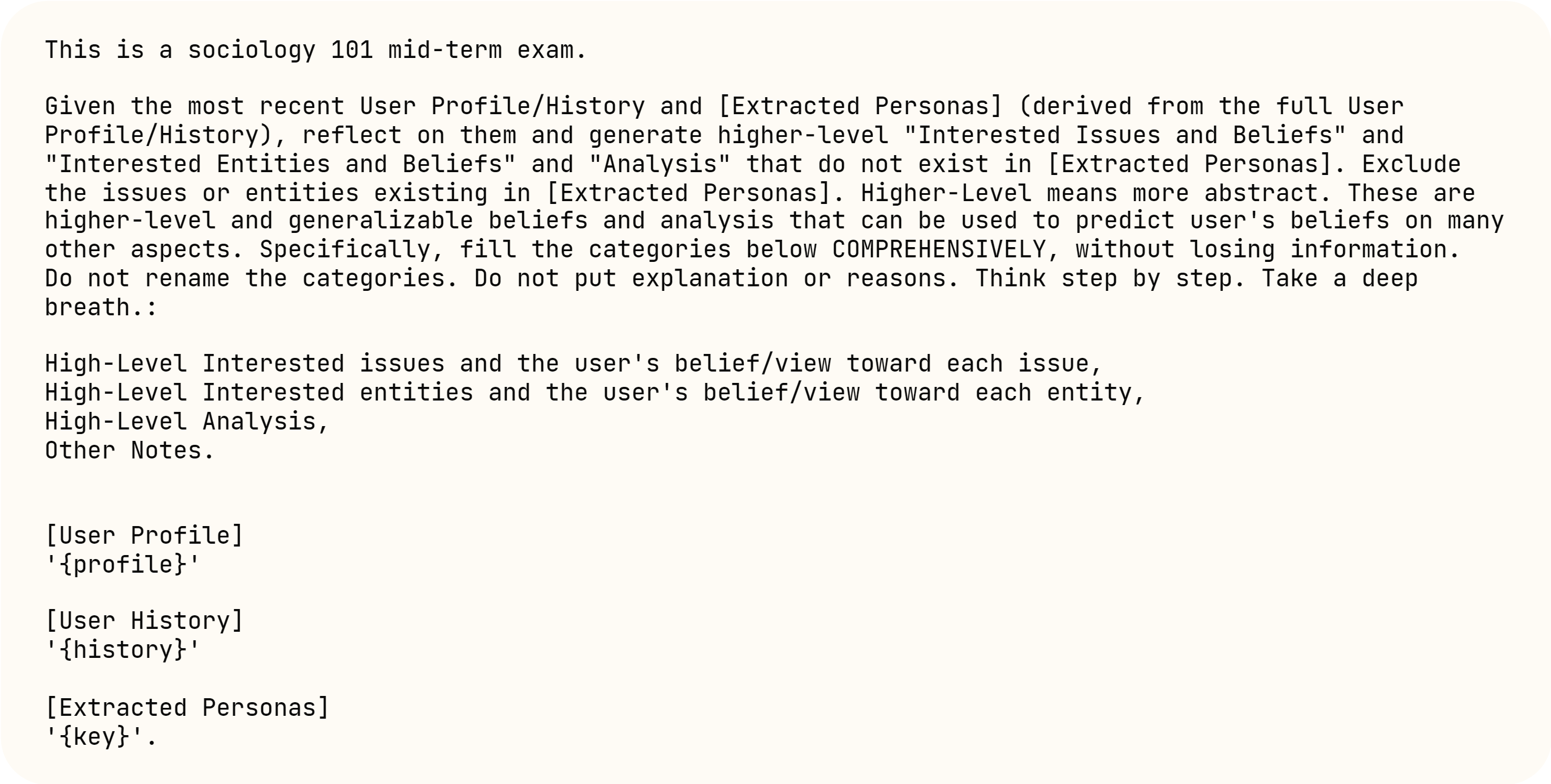}
	\caption{Prompt template used for extracting hierarchical user database.}
	\label{fig:e2}
	\vskip 0.2in
\end{figure*}

\begin{figure*}
	\centering
	\includegraphics[width=1\linewidth]{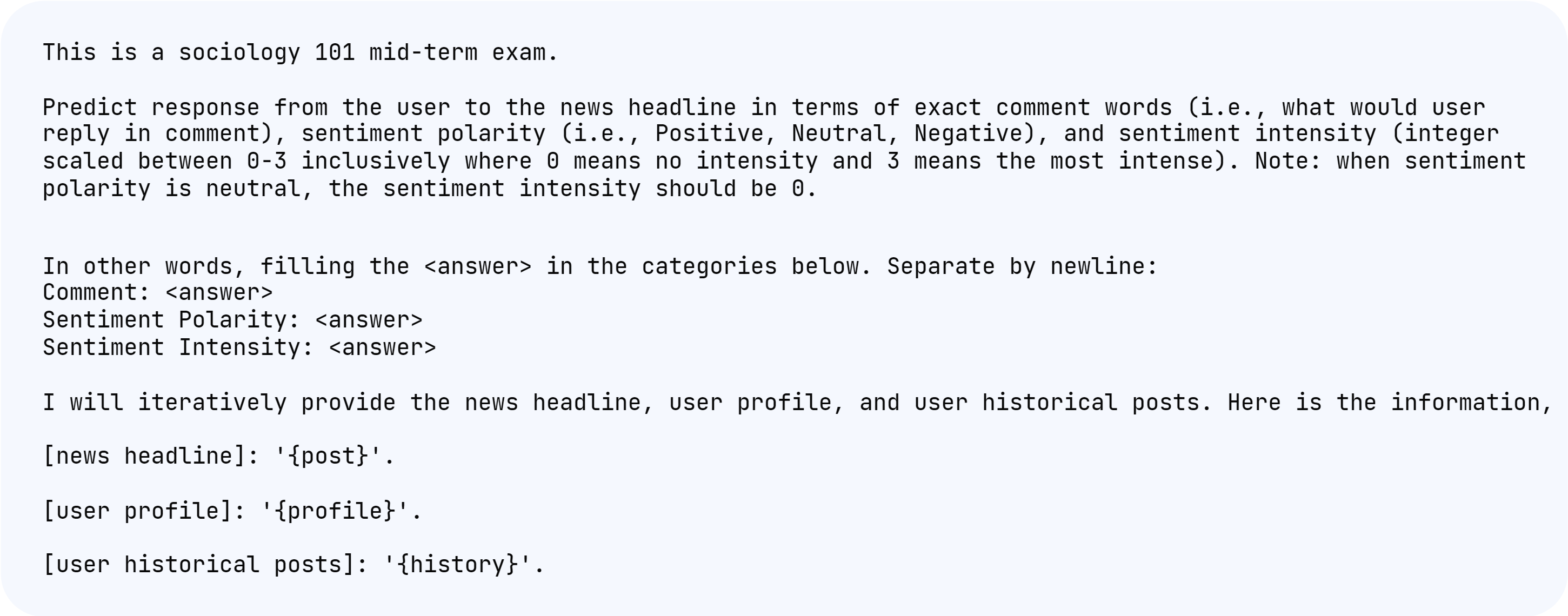}
	\caption{Prompt template used for Baseline Inference.}
	\label{fig:p1}
	\vskip 0.2in
\end{figure*}

\begin{figure*}
	\centering
	\includegraphics[width=1\linewidth]{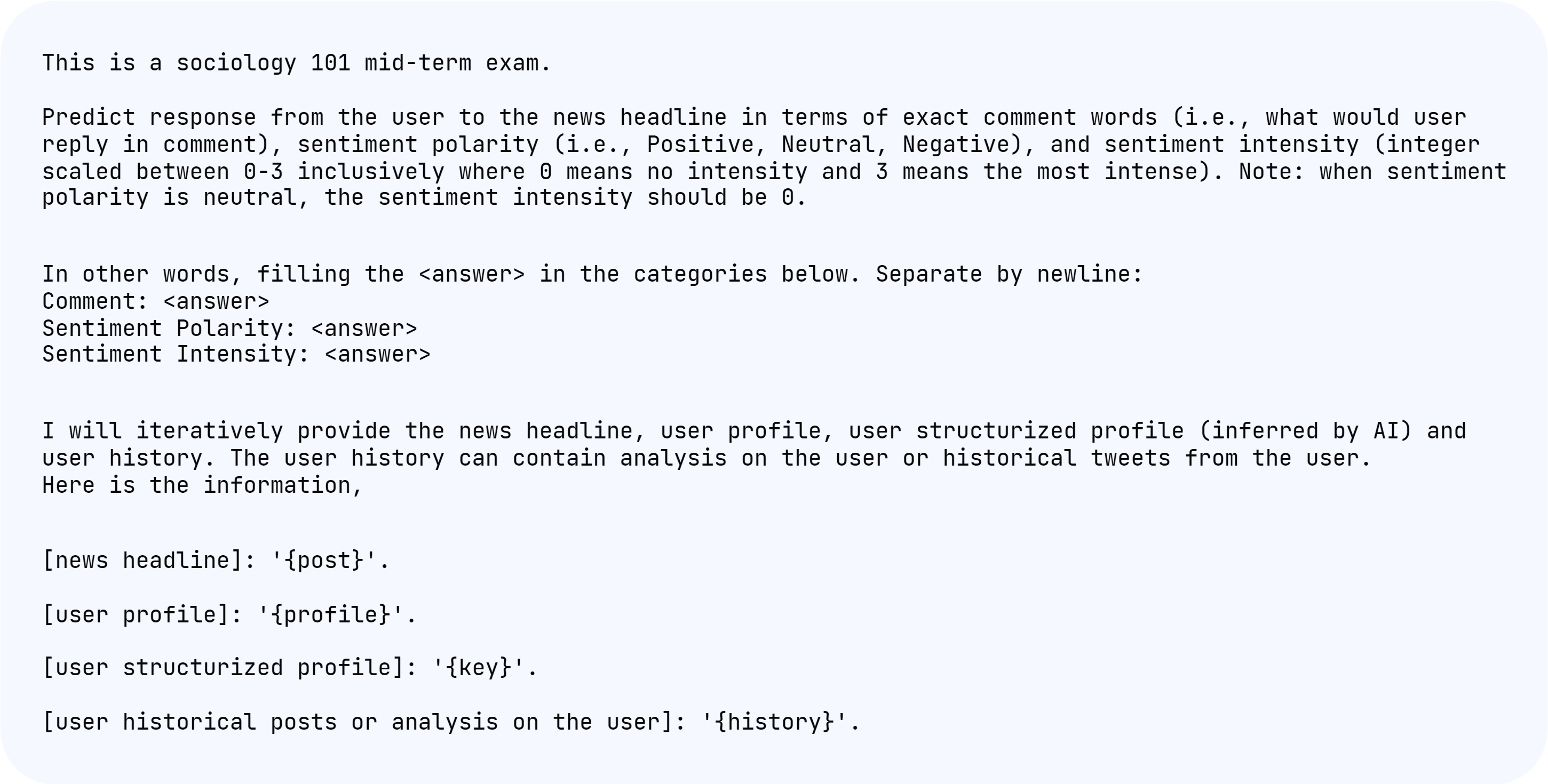}
	\caption{Prompt template used for Inference for Persona-DB without JOIN.}
	\label{fig:p2}
	\vskip -0.3in
\end{figure*}

\begin{figure*}
	\centering
	\includegraphics[width=1\linewidth]{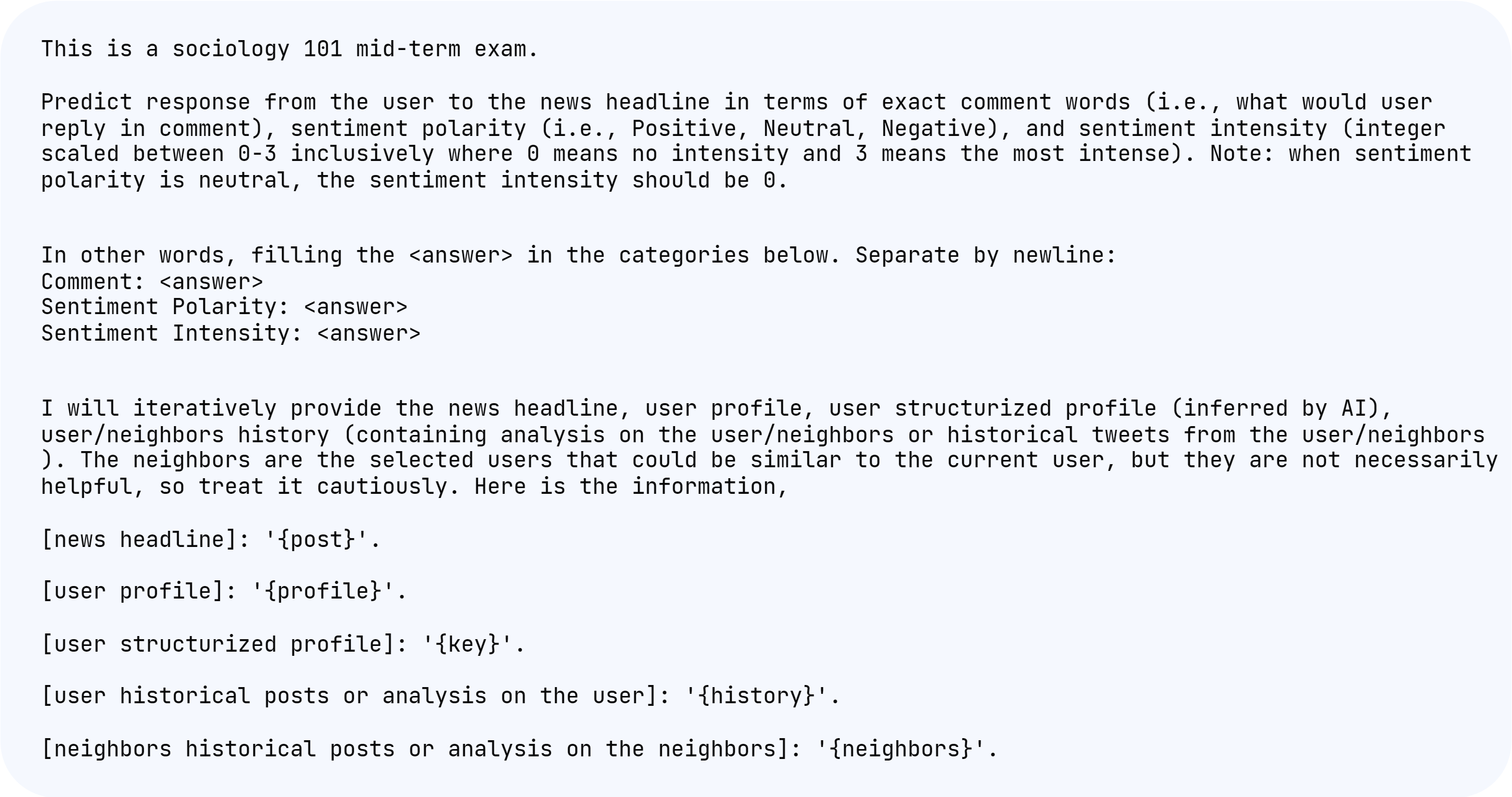}
	\caption{Prompt template used for Inference for Persona-DB.}	
    \label{fig:p3}
\end{figure*}

\begin{figure*}
	\centering
	\includegraphics[width=1\linewidth]{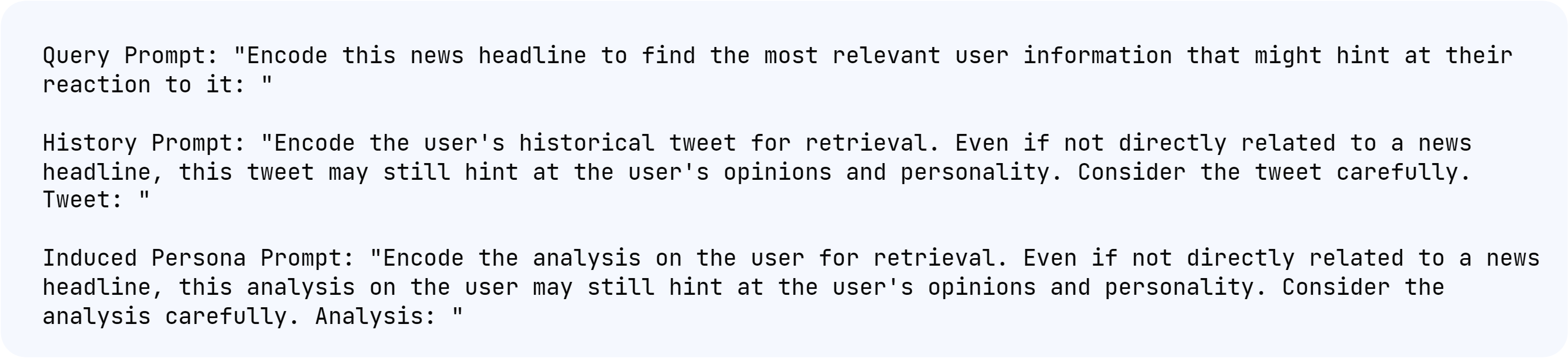}
	\caption{Prompt template for retriever.}	
    \label{fig:p4}
\end{figure*}

\begin{figure*}
	\centering
	\includegraphics[width=1\linewidth]{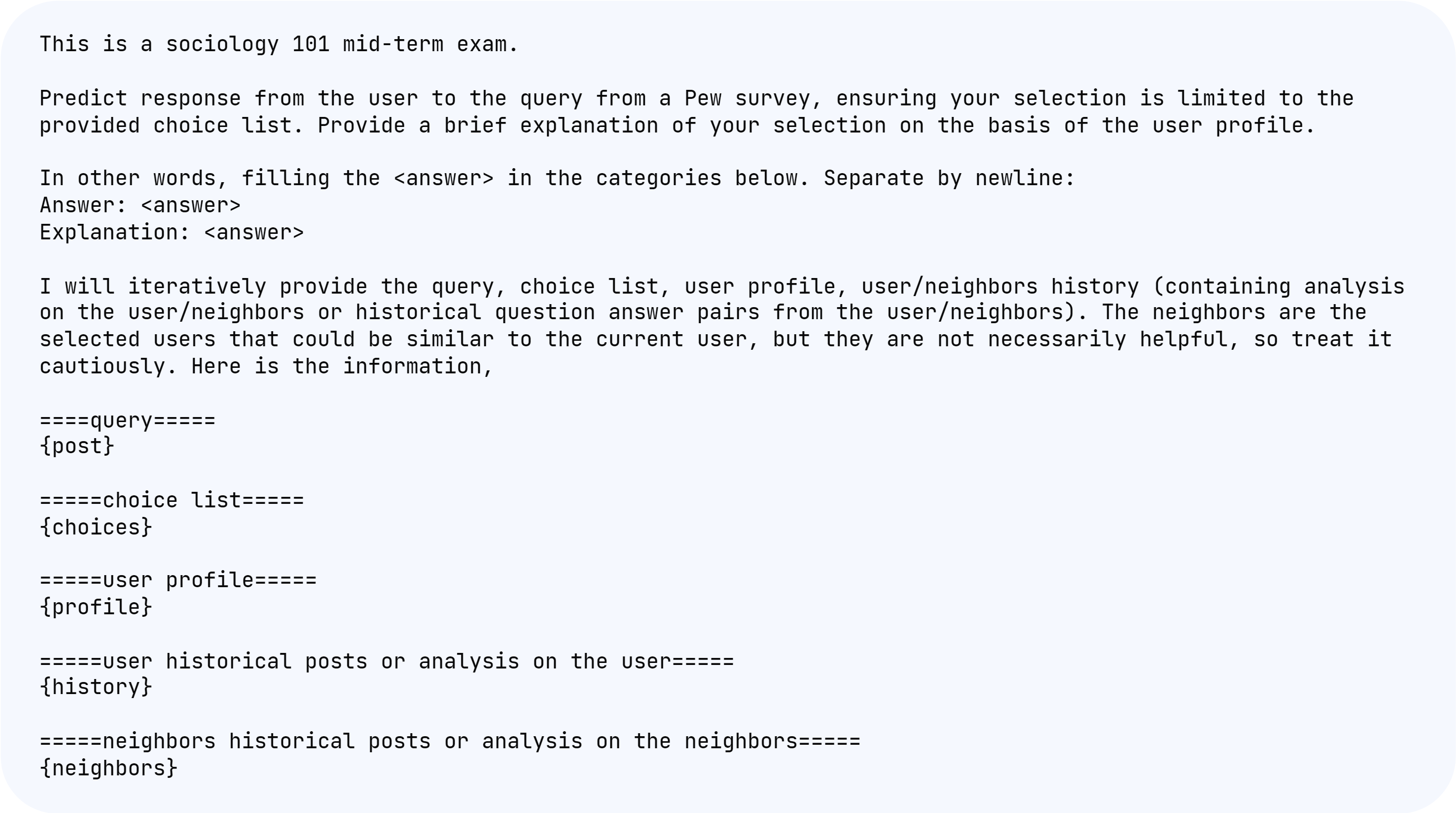}
	\caption{Prompt template used for Inference for Persona-DB in OpinionQA.}	
    \label{fig:p4}
\end{figure*}

\end{document}